\documentclass{article} \usepackage{iclr2027_conference,times}
\iclrpreprintcopy

\usepackage{amsmath,amsfonts,bm}

\def\eqref#1{equation~\ref{#1}}
\def\Eqref#1{Equation~\ref{#1}}

\def\1{\bm{1}}

\DeclareMathAlphabet{\mathsfit}{\encodingdefault}{\sfdefault}{m}{sl}
\SetMathAlphabet{\mathsfit}{bold}{\encodingdefault}{\sfdefault}{bx}{n}

\DeclareMathOperator*{\argmin}{arg\,min}

\usepackage{hyperref}
\hypersetup{hidelinks}
\usepackage{url}
\usepackage{graphicx}
\usepackage{booktabs}
\usepackage[nohyperlinks]{acronym}
\usepackage{amsmath}
\usepackage{amsthm}
\usepackage{enumitem}
\usepackage{multirow}
\usepackage{siunitx}
\usepackage{subcaption}
\usepackage{float}
\usepackage[dvipsnames]{xcolor}
\definecolor{attract1}{rgb}{0.31, 0.53, 0.65}
\definecolor{attract2}{rgb}{0.37, 0.37, 0.68}
\definecolor{reg}{rgb}{0.61, 0.26, 0.37}
\definecolor{hyper}{RGB}{111, 176, 126}

\newtheorem{theorem}{Theorem}
\newtheorem{lemma}{Lemma}
\newtheorem{corollary}{Corollary}[theorem]
\DeclareMathOperator{\tr}{tr}
\DeclareMathOperator{\cov}{Cov}

\acrodef{SSL}[SSL]{self-supervised learning}
\acrodef{VICReg}[VICReg]{variance-invariance-covariance regularization}
\acrodef{SIGReg}[SIGReg]{Sketched Isotropic Gaussian Regularization}
\acrodef{ALLO}[ALLO]{augmented Lagrangian Laplacian objective}
\acrodef{CEM}[CEM]{cross-entropy method}
\acrodef{GB-Planning}[GB-Planning]{gradient-based planning}
\acrodef{CTWM}[CTWM]{Commute-Time-Preserving World Model}
\acrodef{LeWM}[LeWM]{Le World Model}
\acrodef{RL}[RL]{reinforcement learning}
\acrodef{MPC}[MPC]{model predictive control}
\acrodef{GCRL}[GCRL]{goal-conditioned reinforcement learning}
\acrodef{MDP}[MDP]{Markov Decision Process}
\acrodef{logdet}[log-det]{log-determinant}
\acrodef{JEPA}[JEPA]{joint-embedding predictive architecture}

\title{Learning Commute-Time-Preserving World Models for Planning}

\author{Michael Hauri\textsuperscript{*,1,2}, Peter Buttaroni\textsuperscript{*,1,2}, Fabian A. Mikulasch \textsuperscript{1}\& Friedemann Zenke \textsuperscript{1,2} \\
\textsuperscript{1}Friedrich Miescher Institute for Biomedical Research, Basel, Switzerland \\ 
\textsuperscript{2}Faculty of Science, University of Basel, 
Switzerland\\ 
\texttt{\{firstname.lastname\}@fmi.ch} 
}

\begin{document}

\maketitle

\newcommand{\eqcontrib}{\renewcommand{\thefootnote}{*}\footnotetext{Equal contribution.}\renewcommand{\thefootnote}{\dag}\footnotetext{Project page: \url{https://CTWM-website.github.io}}\renewcommand{\thefootnote}{\arabic{footnote}}}
\ificlrfinal
  \eqcontrib
\else
  \ificlrpreprint
    \eqcontrib
  \fi
\fi

\begin{abstract}
World models allow agents to plan in latent space by choosing a sequence of actions that most reduces the distance to a given goal state.
Thus, planning can benefit from latent representations whose distances mirror commute-times in the environment. 
The spectral embedding space of the graph Laplacian provides such a representation, if it obeys a specific eigenvalue-dependent scaling.
Unfortunately, instantiating the graph Laplacian is intractable in large, continuous environments,
and learning its embedding from high-dimensional observations such as raw pixels is difficult.
\Acl{SSL} offers a natural route to such commute-time-preserving embeddings at scale.
However, here we show that existing methods, which commonly encourage isotropic representations to prevent representational collapse, tend to degrade the ``correct'' eigenvalue-dependent scaling, leading to an inaccurate representation of commute times. 
To address this problem, we introduce \acp{CTWM}, a \acs{JEPA}-style world model combining a latent displacement predictor and a log-determinant regularizer that prevents collapse, which provably recover the correctly scaled Laplacian representation under deterministic dynamics and at the predictor's fixed point.
In numerical simulations, \ac{CTWM} matches or outperforms LeWM, a task-agnostic baseline, on several complex, continuous goal-reaching benchmarks, while using half the parameters.$^\dag$
\end{abstract}

\acresetall

\section{Introduction}

Acting intelligently in complex, high-dimensional environments requires imagining the future.
An agent that can simulate the consequences of its actions before taking them can plan, rather than merely react. 
This is the promise of model-based \ac{RL}, where a learned world model might enable fast \ac{RL} and generalization \citep{sutton1990integrated,
 deisenroth2011pilco, chua2018deep, janner2019trust}. 
Still, the world models behind seminal successes \citep[e.g.,][]{Schrittwieser_2020}  are shaped end-to-end by a single, fixed reward function, and adapt poorly when the objective changes \citep{vanseijen2020locaregretconsistentmetric}. 
Animals do not share this inflexibility. For instance, a rat allowed to explore a maze in the absence of any reward will act as it possesses a full map of the maze the instant a reward is introduced \citep{tolman1930introduction, tolman1948cognitive}.
Such \emph{latent learning} suggests that nature has settled on task-agnostic world
models as a strategy. 
What does it take for artificial agents to do the same, reliably and at scale?

\citet{ha2018world} took the first step by learning representation and dynamics
separately by first training a variational autoencoder first and then fitting a  recurrent dynamics model to it afterward. 
Dreamer and its successors instead train both jointly, but through reconstruction \citep{hafner2020dreamcontrollearningbehaviors, hafner2022masteringataridiscreteworld, hafner2025mastering},  which can bias representations on uninformative nuisance variables required to
redraw the scene pixel-by-pixel \citep{deng2021dreamerproreconstructionfreemodelbasedreinforcement}.
More recently, \citet{maes2026leworldmodel} introduced \ac{LeWM}, a \ac{JEPA} that learns a task-agnostic world model trained jointly, end-to-end, \emph{without} reconstruction, using only a lightweight regularizer to prevent collapse. 
This work provides a strong baseline for learning task-agnostic representations, but collapse-avoidance alone does not naturally lead to representational geometries that are also useful for planning \citep{wang2026temporal, nguyen2026latentgeometrysearchamortizing}. 

What properties should a representation suited for planning have?
In classical \ac{RL} on discrete state spaces there exists a principled choice of representation geometry: the \emph{Laplacian representation}, which embeds states using eigenvectors of the transition graph's Laplacian \citep{mahadevan2007proto}, and under an appropriate eigenvalue-dependent rescaling, recovers the graph's commute-time distance in its Euclidean distances \citep{xiao2003resistance, qiu2007clustering}.
This distance accounts for temporal distances between states, and thus forms a natural basis for planning, which has to take obstacles and other relevant factors of the environment into account. 
Learning this representation directly from data is well understood \citep{wu2018laplacian, wang2021towards, gomez2024proper}, with applications to option discovery \citep{machado2017laplacian} and planning \citep{wang2022reachability}.

What is missing is a bridge between these two lines of work: the question is whether the same reconstruction-free, joint-embedding recipe that makes LeWM scale can also be steered toward a representational geometry that is optimal for planning, a question that has not, to our knowledge, been investigated.
We show that this is possible and introduce \ac{CTWM}. 
Our key contributions are:

\begin{itemize}
	\item We introduce \ac{CTWM}, a reconstruction-free world model that jointly learns an encoder and a latent dynamics model by predicting action-induced latent displacements and applying an entropy-maximizing regularizer.
    \item We prove that, under deterministic and reversible dynamics, \ac{CTWM} converges exactly to the Laplacian representation of the environment's transition graph, correctly scaled such that latent distances equal commute-time distances.
	\item On discrete environments with a known transition graph, we confirm this directly: the learned embeddings' pairwise distances match ground-truth commute times, whereas methods encouraging isotropic representations, e.g., \acs{SIGReg}, do not recover the correct scaling.
	\item Finally, we show that \ac{CTWM} outperforms \ac{LeWM} on half the parameter budget
    across LeWM's own goal-reaching benchmarks and two additional visual benchmarks from OGBench (Scene and Pointmaze), 
    all evaluated from pixel observations with no oracle goal representation, which puts the full burden of representation learning on the encoder.
\end{itemize}

\section{Related Work}

World models support sequential decision-making by letting an agent maintain a belief over hidden states, improve a policy through imagined rollouts rather than environment interaction, and plan directly by simulating candidate action sequences at test time.

\textbf{World models for policy learning.} 
\citet{ha2018world} trained a VAE as an encoder and then a predictive model on its latent space to train an agent entirely in imagination. 
More recently, the Dreamer family brought this idea to scale allowing policy iteration in complex environments \citep{hafner2020dreamcontrollearningbehaviors, hafner2022masteringataridiscreteworld, hafner2025mastering}.
While the above works relied on reconstruction objectives for representation learning, the notion was further extended to the reconstruction-free setting \citep{burchi2024mudreamerlearningpredictiveworld, hauri2026dreamer}. 

\textbf{World models for planning.} 
To plan directly by simulating action sequences, classic work searched decision trees of future moves using known environment dynamics \citep{silver2016mastering}, later extended to learned transition models when the true dynamics are unavailable \citep{Schrittwieser_2020}. 
However, these approaches depend on reward and thus on the policy. 
This dependence typically precludes zero-shot generalization, as it requires retraining the model when the task or the rules of the game change.

\textbf{Task agnostic world models} address this limitation by learning representations and transition models from data without requiring reward.
One line of work first trains  an encoder network on generic image or video data~\citep{zhou2024dino, assran2025v} using reconstruction-free \ac{SSL} and then learns a latent dynamics model on top for planning. 
More recently several studies relied on \acp{JEPA}, allowing the encoder and predictor, i.e., the dynamics model, to be trained jointly  \citep{sobal2026learning, maes2026leworldmodel, zhao2026sub}, thereby further boosting planning performance and enabling zero-shot generalization.
To mitigate the compounding rollout error such joint-embedding predictors accumulate over long horizons, some models predict jumpy, temporally abstract transitions instead \citep{farebrother2025temporal, farebrother2026compositional}.
The main difference of the above reconstruction-free approaches is how they avoid collapse.

\textbf{Avoiding collapse in \ac{SSL}:} To prevent trivial solutions in \ac{SSL} different methods use distinct mechanisms. On the one hand, contrastive methods pull together the representations of positive samples while pushing apart negative ones~\citep{sohn2016improved, oord2018representation, chen2020simple}. 
On the other hand, non-contrastive methods forgo negatives and either rely on asymmetric architectures using a predictor and a stop-gradient, for instance, SimSiam \citep{chen2021exploring} and BYOL \citep{grill2020bootstrap}, or additional regularization (Barlow Twins; \citealp{zbontar2021barlow}) as well as \ac{VICReg}  \citep{bardes2021vicreg}.
More recently \citet{balestriero2025lejepa} proposed \ac{SIGReg} that matches the full projected embedding distribution to an isotropic Gaussian. 
Most existing \ac{SSL} methods encourage isotropic representations either explicitly or implicitly \citep{halvagal_implicit_2023}, which can pose challenges for learning scaled representations.

\textbf{Laplacian representations}  embed states using eigenvectors of the graph Laplacian \citep{belkin2003laplacian} induced by the \ac{MDP}'s transition structure \citep{mahadevan2007proto}, and are closely related to slow feature analysis~\citep{wiskott2002slow, sprekeler2011relation, richthofer2015predictable}. 
In \ac{RL}, \citet{wu2018laplacian} proposed an objective for learning these representations directly from data, which was improved subsequently \citep{wang2021towards}. 
\citet{gomez2024proper} introduced \ac{ALLO}, a method explicitly designed to recover both the Laplacian eigenvectors and eigenvalues from sampled transitions through a constrained optimization scheme.
Laplacian representations have served to define spatio-temporally abstract states~\citep{shehmar2026laplacian} and intrinsic reward functions ~\citep{machado2017laplacian}.
Connections between \ac{SSL} objectives and spectral embedding methods were already studied by \citet{balestriero2022contrastive}.
Although the above methods recover the Laplacian geometry, they neither necessarily learn a latent dynamics model nor guarantee appropriate eigenvalue-dependent scaling.

\textbf{Alternative latent geometries for goal-reaching.} A separate line of work shapes latent representations directly around planning-relevant distances, by learning (quasi-)metric embeddings via temporal-difference estimation of the transit time between states and goals \citep{wang2023optimal, myers2026offline, park2026dual, zheng2025multistep}.
Unlike our method, these models only compute temporal distances between states, but do not train a world model with latent state predictor end-to-end.
A second body of work learns successor features that aggregate discounted future state occupancies~\citep{eysenbach2022contrastive, bagatella2026td}. 
While these methods enable zero-shot goal-reaching, their learned distances are inherently policy-dependent rather than the intrinsic structure of the environment. 
In contrast, \ac{CTWM} recovers the environment's commute-time distance.

\section{Constructing \aclp{CTWM}}

Before we start, we briefly introduce the basic notation. Let $\mathcal{S}$ be the observation space and $\mathcal{A}$ the action space. 

\emph{World model.} An encoder $\phi : \mathcal{S} \to \mathcal{Z}$ maps observations to a latent space $\mathcal{Z}$, and a predictor $T : \mathcal{Z} \times \mathcal{A} \to \mathcal{Z}$ simulates the dynamics in that space, $\hat z' = T(z, a)$, where $z = \phi(s)$.

\emph{Transition graph.} Let $P_\pi$ be the node transition matrix under the behavioral policy $\pi$ and $\rho$ its stationary distribution. 
The transition graph
$G = (V, W_\pi)$ has the symmetrized weights $W_\pi = \tfrac12\bigl(\operatorname{diag}(\rho)\, P_\pi + P_\pi^\top \operatorname{diag}(\rho)\bigr)$, which sum to one, and degree matrix $D = \operatorname{diag}(\rho)$.
Its Laplacian is $L_\pi = D - W_\pi$, and we denote by $(\lambda_k, u_k)_{k \ge 0}$ the solutions of the generalized eigenproblem $L_\pi u_k = \lambda_k D u_k$, normalized as $u_k^\top D u_\ell = \delta_{k\ell}$.

\subsection{Planning in metric latent spaces}
\label{sec:planning}
Planning refers to finding the action sequence that minimizes a cost $J$ over a horizon $H$. 
This cost can be reward-based, as in \citet{hansen2024td}, or reward-free, as is common in \acl{GCRL}. 
In the latter case, the cost measures the $L_2$ distance between the final latent state and an encoded goal $z_g = \phi(s_g)$:
\begin{equation}
\label{eq:planning}
    a_{1:H}^\star = \argmin_{a_{1:H}} J(z_H, z_g)
    = \argmin_{a_{1:H}} \lVert z_H - z_g \rVert_2 \quad ,
\end{equation}
where $z_H$ is obtained by recursively applying the learned dynamics from $z_0 = \phi(s_0)$. 
This problem is commonly solved with the sampling-based \ac{CEM} \citep{rubinstein2004cross}, or gradient-based methods \citep{s2023gradientbasedplanningworldmodels}. 
Common to both approaches is that they rely on computing distances in the latent space, which requires these distances to be meaningful in the environment. 
A prominent choice is to require points in the latent space to preserve commute-time distances in the environment \citep{xiao2003resistance, qiu2007clustering}.
On a graph, the commute time $C(s,s^\prime)$ corresponds to the expected number of steps for a random walk to travel from node $s$ to node $s'$ and back. 
It relates to the eigenvalues and eigenvectors $(\lambda_k, u_k)_{k \ge 1}$ of the graph Laplacian through
\begin{equation}
    C(s,s') = \sum_{k \ge 1}
    \frac{\bigl(u_k(s) - u_k(s')\bigr)^2}{\lambda_k} \quad .
\end{equation}
For latent distances to correspond to commute time distance, we further define the scaled Laplacian representation \citep{qiu2007clustering, wang2022reachability, shehmar2026laplacian} as $\psi_k(s) := \lambda_k^{-1/2} u_k(s)$, which yields 
\begin{equation}
    C(s,s') = 
    \lVert \psi(s) - \psi(s') \rVert_2^2 \quad ,
\end{equation}
so that squared $L_2$ distances in $\psi$ are commute-time distances.

In other words, there exists a representation space in which the commute-time distance is Euclidean.
It requires an eigenvalue-dependent scaling of the representations, which is realized by the scaled Laplacian representation.
This property makes the scaled Laplacian representation well-suited for planning and thus a desirable learning target for \ac{SSL} algorithms whose embeddings should facilitate planning  in downstream tasks.
Crucially, the scaled Laplacian representation is anisotropic for most environments with non-degenerate eigenvalues.
In contrast, existing \ac{SSL} methods commonly encourage isotropic embeddings.
Before we can study whether and to what extent different \ac{SSL} methods recover the scaled Laplacian representation, we first present the general framework for learning task-agnostic world model with \ac{SSL}.

\subsection{Learning task-agnostic world models with \ac{SSL}}
Given a task-agnostic dataset of transitions $\{(s_k, a_k, s'_k)\}_{k=1}^N$, collected without rewards or task labels by an arbitrary behavior policy, the canonical approach to learning a world model via \ac{SSL} is specified by the tuple $(\phi, T)$ described above. 
Both components are typically learned by minimizing
\begin{equation}
    \mathcal{L}(\phi, T) = \mathbb{E}_{(s,a,s')}\left[\,\lVert T(z, a)
    - z' \rVert^2\,\right] + \mathrm{Reg}(\phi) \quad ,
    \label{eq:vanilla_jepa}
\end{equation}
where $\mathrm{Reg}(\phi)$ denotes a regularizer that prevents representational collapse.
For instance, LeWM \citep{maes2026leworldmodel} relies on \ac{SIGReg} \citep{balestriero2025lejepa}, but other regularizers are possible. 
Importantly, the regularizer may also be realized implicitly, e.g., through a stop-gradient operation \citep{grill2020bootstrap, tian_understanding_2021, halvagal_implicit_2023}, as long as it preserves representational entropy \citep{shwartz2023information, mikulasch2026understanding}.

Our \ac{CTWM} approach deviates from the formulation above in two distinct ways:
We use residual prediction and entropy regularization to prevent collapse, which we will explain in the following.

\subsection{Learning world models via residual predictions}
\begin{figure}[tb]
    \centering
    \includegraphics[width=1.0\linewidth]{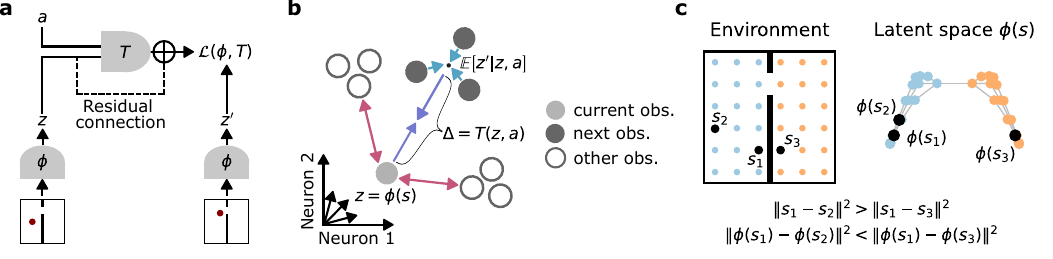}
    \caption{ \textbf{(a)}~Overview of CTWM: a shared encoder $\phi$ maps observations $s$ and $s'$ to the representations $z$ and $z'$ respectively. 
    The transition model $T$ conditioned on action $a$, predicts a residual displacement $\Delta$ that is added to the current embedding via a residual connection. 
    The loss $\mathcal{L}(\phi, T)$ is computed from the predicted and actual next-state embeddings.
    \textbf{(b)}~Geometric interpretation of the loss $\mathcal{L}(\phi,T^\star)$ in the latent space; arrow colors refer to terms in \eqref{eq:trcov}. 
    The loss terms encourage predictable transitions (teal), penalize large distances between consecutive points (blue), and prevent collapse of representation (magenta).
    \textbf{(c)}~Latent distances reflect commute-time distance rather than spatial proximity. Left: Two-room connected by a door; $s_1$ and $s_3$ are adjacent but separated by a wall (dashed line). Right: in the \ac{CTWM} embedding the rooms unfold around the door, placing $s_1$ near $s_2$ and far from $s_3$.
    } 
    \label{fig:Fig_1}
\end{figure}

The first difference of \ac{CTWM} to existing approaches such as \ac{LeWM} is residual prediction.
Concretely, the predictor $T$ is trained to predict the residual or displacement between successive embeddings $\Delta(z,z') = z'-z$ such that the next embedding is given by $\hat{z}' = z + T(z,a)$ (Figure~\ref{fig:Fig_1}a). 
The encoder and displacement are then trained jointly by minimizing:
\begin{equation}
\label{equ:CTWM}
    \mathcal{L}(\phi,T) = \mathbb{E}_{(s,a,s')}[\lVert\Delta(z,z') - T(z,a)\rVert^2 + \beta\lVert T(z,a)\rVert^2] + Reg(\phi) \quad ,
\end{equation}
where the $\beta\lVert T(z,a)\rVert^2$ term acts as a regularizer that penalizes large changes \citep[][]{saanum2024simplifying}, which can also be seen as a slowness objective \citep{wiskott2002slow, sprekeler2011relation}. 
Without this term, a sufficiently expressive predictor can absorb arbitrarily large latent displacements, leaving the encoder free to map temporally consecutive states to distant points in the latent space. 
By penalizing large predictor outputs, $\beta > 0$ forces the encoder to keep consecutive latent states close together: the only way to simultaneously achieve low prediction error and low predictor norm is to produce small displacements. 
Where required, the above loss can be equivalently written as a next-state prediction problem instead of residual prediction,  by simply adapting the regularizer (see Appendix~\ref{app:smoothness}). 

For deterministic dynamics, one can show that the embeddings $z_t$ resulting from optimizing \eqref{equ:CTWM} converge to the Laplacian subspace if the regularization $Reg(\phi)$ introduces an orthonormality constraint (Appendix~\ref{app:conv}).
To understand this intuitively, it is helpful to assume that the predictor is at its fixed point of learning $T^\star = \frac{\mathbb{E}[\Delta|z,a]}{1+\beta}$.
This assumption allows us rewrite the loss as follows
\begin{equation}
\label{eq:trcov}
    \mathcal{L}(\phi,T^\star) =
     {\color{attract1} \mathbb{E}_{(s,a)}[\tr\bigl(\cov(\Delta \mid z,a)\bigr)] }+ 
      {\color{attract2} \frac{\beta}{1+\beta}\mathbb{E}_{(s,a)}[\lVert\mu(z,a)\rVert^2] }
      +  {\color{reg} Reg(\phi)} \quad ,
\end{equation}
where $\mu(z, a) = \mathbb{E}[\Delta|z,a]$ is the mean conditioned displacement and the colors correspond to the diagram in Figure~\ref{fig:Fig_1}b.
Given a stochastic transition function $p(s'|s,a)$, the first term tries to make $p(z'|z,a)$ predictable by making the sum of variances small, all the while
the second term attracts states that are correlated in time together. 
Finally, the last term prevents global collapse.
The fixed point condition  also implies that we should use $(1+\beta)T$ for rollouts instead of $T$ (Appendix \ref{app:beta}).

\subsection{Regularization to prevent collapse}
\label{sec:regularizers}
The second distinction of \ac{CTWM} from previous work is that it uses 
entropy regularization to prevent collapse.
Specifically, we use the \ac{logdet} of the covariance matrix, which corresponds to the entropy of the multivariate Gaussian, as the regularizer $Reg(\phi)$. 
The underlying reason is that different regularizers converge to different representations.
For example, a hard orthonormality constraint (Appendix \ref{app:ortho}) yields a Laplacian representation, but without eigenvalue-dependent scaling by design \cite[][Appendix \ref{app:scalings}]{wu2018laplacian}. 
The constraint can be relaxed through VCReg \citep{zhu2023variance}, which enforces a minimal floor on the diagonal of the covariance matrix  (Appendix~\ref{app:vcreg}).
Assuming that this regularizer converges to the Laplacian representation, the eigenvalue-dependent scaling is constant for the smallest eigenvalues and $\propto \lambda_k^{-1}$ for the rest (Appendix~\ref{app:scalings}). 
Thus many regularizers used in \ac{SSL} encourage an isotropic covariance spectrum, either explicitly \citep{zhu2023variance, zbontar2021barlow, balestriero2025lejepa} or implicitly \citep{halvagal_implicit_2023}.
Generally this isotropy is incompatible with the scaled Laplacian representations and, thus, does not represent commute-times accurately. 

To address this issue, we use \ac{logdet} regularization, which instead constrains the product of the eigenvalues.
Crucially, it does not impose a hard threshold on feature variance and VCReg can be seen as an approximation when the representational covariance matrix is close to diagonal \citep{shwartz2023information,mikulasch2026understanding}.
We will show in the next section that \ac{logdet} gives representations enough flexibility to achieve $\lambda_k^{-1/2}$ scaling, as required for capturing commute-time distances.

\subsection{Convergence to the scaled Laplacian representation}
\label{sec:theory}

Combining residual prediction with \ac{logdet} regularization yields the \ac{CTWM} training objective:
\begin{equation}
\label{equ:CTWM_full}
\boxed{
    \mathcal{L}(\phi, T)
    = 
    \mathbb{E}_{(s,a,s')}\bigl[
    {
        \lVert \Delta(z,z') - T(z, a) \rVert^2}
      + 
      { \beta \lVert T(z, a) \rVert^2
      }
      \bigr]
    -  { \frac{\gamma}{2} \log \lvert \Sigma \rvert} ~ ,
    }
\end{equation}
with the two hyperparameters $\beta$ and $\gamma$ and the covariance matrix $\Sigma \equiv \operatorname{Cov}_\rho\bigl(\phi(s)\bigr)$ computed over the stationary transition distribution $\rho$. 
In the following we will show that this objective is well suited for learning world models for planning. 
Our main result is that, at the optimum, the encoder recovers the scaled Laplacian representation.

\begin{theorem}[Recovery of scaled Laplacian representation]
\label{thm:scaled_lap}Let $(\lambda_k, u_k)_{k \ge 0}$ be the eigenvalues and eigenvectors
of the generalized eigenproblem of $L_\pi$, ordered as $0 = \lambda_0 < \lambda_1 \le \dots \le \lambda_{n-1}$
with $\lambda_d < \lambda_{d+1}$, and let
\begin{equation}
\label{eq:scaled_lap}
  \psi(s) = \bigl(\lambda_k^{-1/2} u_k(s)\bigr)_{k=1}^{d},
  \qquad d = \dim \mathcal{Z},
\end{equation}
be the scaled Laplacian representation. Under deterministic dynamics, and 
assuming the predictor is at its fixed point $T^\star = \frac{\mathbb{E}[\Delta|z,a]}{1+\beta}$,  equation~\ref{equ:CTWM_full} becomes: 
\begin{equation}
\label{eq:deter}
  \mathcal{L}(\phi, T^\star)
  = \frac{\beta}{1+\beta}\,
    \mathbb{E}_{(s,s')}\bigl[\lVert \phi(s') - \phi(s) \rVert^2\bigr]
  - \frac{\gamma}{2} \log \lvert \Sigma \rvert,
\end{equation}
and is minimized, up to a rotation and a translation, by
\begin{equation}
\label{eq:opt_phi}
  \phi^\star(s) = \sqrt{\tfrac{\gamma(1+\beta)}{4\beta}}\; \psi(s) \quad .
\end{equation}
\end{theorem}
The proof is given in Appendix~\ref{app:th1_proof}.
The eigendirections are the slowest ones of the Laplacian, so they are shared with VCReg \citep{balestriero2022contrastive} and orthonormality constraints \citep{wu2018laplacian}, but what distinguishes the \ac{logdet} from the two other regularizers is the eigenvalue-dependent scaling $\lambda_k^{-1/2}$, which carries the metric information relevant for planning.
This relationship between the commute-time distance and the $L_2$ distance in the learned space can be made explicit as follows:
\begin{corollary}[Relationship to commute-time distance]
\label{cor:ctd}
Let
\begin{equation}
\label{eq:ctd}
  C_d(s,s') := \sum_{k=1}^{d}
  \frac{\bigl(u_k(s) - u_k(s')\bigr)^2}{\lambda_k}
\end{equation} 
be the rank-$d$ truncation of the commute-time distance on the transition graph \citep{shehmar2026laplacian},
then
\begin{equation}
  \bigl\lVert \phi^\star(s) - \phi^\star(s') \bigr\rVert_2^2
  = \frac{\gamma(1+\beta)}{4\beta}\, C_d(s,s')
  \qquad \text{for all } s, s' \in \mathcal{S}.
\end{equation}
\end{corollary}
This result follows directly from combining equation~\ref{eq:opt_phi} with equation~\ref{eq:ctd}. 
To verify this relation experimentally, we further link the eigenvalues of the underlying graph to the variance of the representations:
\begin{corollary}[Variance of the representation]
\label{cor:var}
The $k$-th eigendirection of $\phi^\star$ carries variance
\begin{equation}
  \operatorname{Var}\bigl(\phi_k^\star\bigr(s))
  = \frac{\gamma(1+\beta)}{4\beta\,\lambda_k} \quad.
\end{equation}
\end{corollary}
In other words, slow directions are amplified and fast ones inhibited (see Appendix~\ref{app:corollary} for the proof). 
For comparison we compute the ratios $\operatorname{Var}(\phi_k(s))/\operatorname{Var}(\phi_1(s))$, which are equal to $\lambda_1/\lambda_k$ for our loss, to get rid of the global scale factor.
Together, these results show that regularizers for preventing representational collapse may result in differently scaled representations.
Specifically, the analysis shows that orthonormality constraints and VCReg are not consistent with the scaled Laplacian representation, whereas the \ac{logdet} supports a correctly scaled representation, with variance $\propto \lambda_k^{-1}$ and $L_2$ distances equal to commute times (Figure~\ref{fig:Fig_1}c). 
In the following, we will confirm these findings in numerical experiments.

\section{Experiments}
\begin{figure}[tb]
    \centering
    \includegraphics[width=1.0\linewidth]{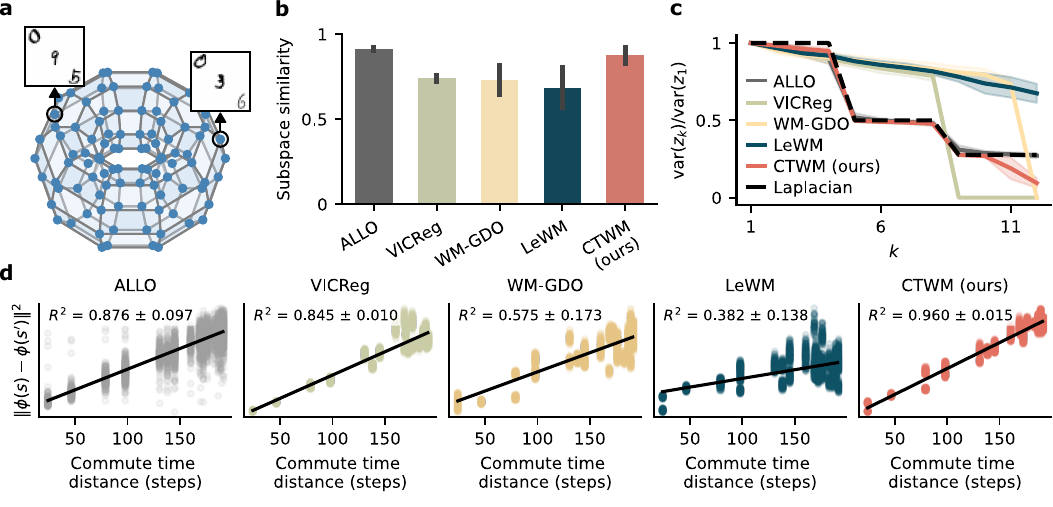} 
    \caption{\textbf{Torus environment and empirical verification of the theoretical predictions.} 
    \textbf{(a)}~The torus graph: each vertex carries a unique three-digit label, randomly shuffled across the graph, observed as three MNIST images sampled independently from the corresponding classes at each visit.
    \textbf{(b)}~Subspace similarity to the ground-truth Laplacian eigenbasis for the five methods $\pm$ SD ($n=4$; see Appendix~\ref{app:subsim}).
    \textbf{(c)}~Variance ratio $\text{var}(z_k)/\text{var}(z_1)$ against eigenvalue index $k$ (cf.\ Corollary~\ref{cor:var}), with the ground-truth scaling $\lambda_1/\lambda_k$ in dashed black. Only \ac{ALLO} and \ac{CTWM} track it, confirming that the \ac{logdet} regularizer recovers the predicted scaling, $\text{var}(z_k) \propto \lambda_k^{-1}$; \ac{VICReg}, WM-GDO and \ac{LeWM} show flatter spectra, with \ac{VICReg} and WM-GDO collapsing the trailing dimensions.
    \textbf{(d)}~Squared $L_2$ latent distance versus ground-truth commute-time distance, one point per pair of graph nodes. Black line: least-squares linear fit, with its coefficient of determination $R^2$. \ac{CTWM} achieves the highest $R^2$. $R^2$ reported as mean $\pm$ SD over $n = 4$ seeds.} 
    \vspace{-3mm}
    \label{fig:torus}
\end{figure}

We first sought to verify that \ac{CTWM} learns successfully in a setting in which the ground-truth Laplacian representation  can be computed explicitly. 
To this end, we constructed a simple environment in which the agent moves along the edges of a discrete toroidal graph (Fig.~\ref{fig:torus}a). 
Each node on the graph corresponded to an image displaying a unique three digit sequence; sequences were randomly assigned to nodes. 
To make this a non-trivial representation learning problem,
the digits' images were drawn from the MNIST dataset \citep{lecun1998gradient} and every time the agent revisited a given state, a different digit corresponding to the same class was drawn. 
The underlying graph dynamics are thus deterministic, while the observations, and hence the transitions in observation space, are stochastic, requiring the encoder to learn digit representations that generalize.

We trained a small \ac{CTWM} model on 20,000 transitions in the above environment using a random policy. 
The encoder was a small convolutional network that produced a 12-dimensional embedding, and the predictor was a single-hidden-layer MLP with 32~units. 
The action was one-hot encoded and we set the displacement penalty to $\beta=0.1$. 
All simulations were run for $n=4$ seeds. 

To check whether \ac{CTWM} had learned the correct representational subspace shared with the actual spectral embedding of the graph Laplacian and the correct eigenvalue scaling, we computed the subspace similarity of the learned embeddings to the analytical solution.
For comparison, we further computed the subspace similarity for \ac{VICReg} and \ac{LeWM}, as representative examples of non-residual world-modeling approaches; \ac{ALLO} \citep{gomez2024proper}, which served as a proxy of the Laplacian representation; and WM-GDO, which keeps \ac{CTWM}'s residual prediction but replaces the log-det regularizer with a graph-drawing orthonormality constraint \citep{koren2003spectral,wu2018laplacian}.

We found that \ac{ALLO} and \ac{CTWM} recovered the Laplacian subspace most closely (subspace similarity $\approx$ 0.9), whereas \ac{VICReg}, WM-GDO and \ac{LeWM} achieved lower alignment, with WM-GDO and \ac{LeWM} also being more variable across seeds (Fig.~\ref{fig:torus}b).

Subspace similarity, however, is invariant to the scaling of the embedding. We therefore examined the variance along each eigendirection: only CTWM and ALLO reproduced the ground-truth $\lambda_1/\lambda_k$ staircase, whereas \ac{VICReg}, WM-GDO and \ac{LeWM} displayed flatter spectra, with \ac{VICReg} and WM-GDO additionally collapsing the trailing dimensions (Fig.~\ref{fig:torus}c). Finally, these differences were reflected in how faithfully the embeddings encode commute times: \ac{CTWM} achieved the highest $R^2$ ($0.960 \pm 0.015$; Fig.~\ref{fig:torus}d), followed by \ac{ALLO} ($0.876 \pm 0.097$) and \ac{VICReg} ($0.845 \pm 0.010$), while WM-GDO ($0.575 \pm 0.173$) and \ac{LeWM} ($0.382 \pm 0.138$) captured commute-time geometry only poorly. Thus, \ac{CTWM} closely recovers the Laplacian spectral subspace and its eigenvalue scaling, and learns embeddings whose distances most faithfully encode commute times.

\subsection{Learning CTWM in continuous RL environments}
To evaluate whether the advantage of representing commute-time geometry transfers beyond discrete graphs, we compared \ac{CTWM} against \ac{LeWM} \citep{maes2026leworldmodel} on the following six benchmark tasks. For (i--iv) we used the dataset provided by \citet{maes2026stable}, while for (v, vi) we used two environments from \citep{park2025ogbench}. 
All environments provide pixel observations.
(i)~\emph{Two-Room}: 2D navigation between two rooms joined by a door~\citep{sobal2026learning}, with trajectories from a noisy heuristic policy.
(ii)~\emph{Reacher}: a two-link arm reaching a target point in the plane~\citep{tassa2018deepmind}, with trajectories from a trained soft actor-critic policy.
(iii)~\emph{PushT}: a point agent pushing a T-shaped object to a target pose~\citep{zhou2024dino}, with expert demonstrations.
(iv)~\emph{Cube}: 3D manipulation in which a robot arm moves a cube between positions, with trajectories from the heuristic policy of the benchmark~\citep{park2025ogbench}.
(v)~\emph{Pointmaze}: 2D maze navigation, using the 1000 trajectories of the OGBench medium-navigate dataset~\citep{park2025ogbench} converted to visual inputs (Figure~\ref{fig:pointmaze}a).
(vi)~\emph{Scene}: 3D manipulation with diverse objects (drawer, window, cube, buttons)~\citep{park2025ogbench}, with data from~\citet{maes2026stable}.

As in \citet{maes2026leworldmodel}, we predicted 25~steps ahead with a maximum planning budget of 50~steps.
To implement \ac{CTWM}, we replaced \ac{LeWM}'s \ac{SIGReg} training objective with \eqref{equ:CTWM_full}, while using the same ViT-T encoder with 5.5M parameters. 
For the predictor, we retained the deep-AdaLN architecture with six layers but reduced the latent dimension from 192 to 64, which decreased the predictor's parameter count from 11M to 3M.
The total model size was 9M parameters, compared to 18M for LeWM.

\subsubsection{CTWM results in superior planning performance}
\begin{figure}[tb]
    \centering
    \includegraphics[width=1.0\linewidth]{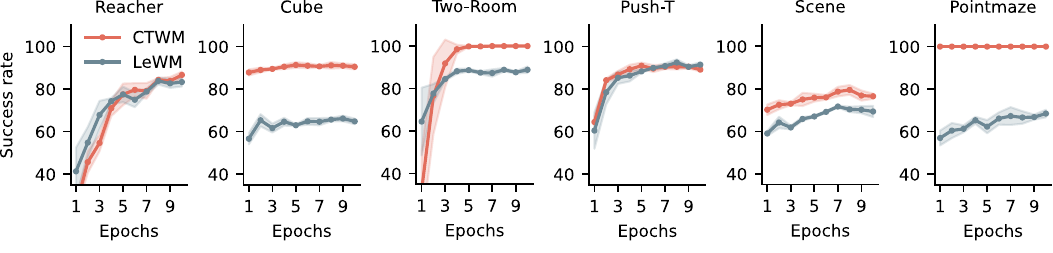}
    \caption{\textbf{Success rates of \ac{CTWM} and \ac{LeWM} on the six benchmark tasks:} Reacher, Cube, Two-Room, PushT, Scene and Pointmaze. \ac{CTWM} (red) matches or outperforms \ac{LeWM} (blue) on every environment using half the number of parameters. Shaded regions indicate the SD ($n=3$).} 
    \label{fig:CTWM_lewm_perf}
\end{figure}
Following the steps of \citet{maes2026leworldmodel}, we  evaluated planning accuracy using  \ac{CEM} on models
trained with \ac{LeWM}, \ac{LeWM}+TS (temporal straightening; \citealp{wang2026temporal}), and \ac{CTWM} on all six environments for $n=3$ seeds each, using the same hyperparameters across all of them. 
\ac{CTWM} performed either better than, or on par with LeWM on all tasks (Fig.~\ref{fig:CTWM_lewm_perf}; Table~\ref{app_table:numerical_results}).
On Two-Room, where \cite{maes2026leworldmodel} noted that \ac{SIGReg} can be detrimental in low-complexity environments, \ac{CTWM} reached $100\%$ success rate, matching the ceiling obtained by other methods. 
On Reacher and PushT, both methods achieved comparable performance. 
The clearest gains appeared on the three harder tasks: Cube, Scene, and Pointmaze, on which \ac{CTWM} consistently outperformed \ac{LeWM}, with larger margins during early training. 
Notably \ac{CTWM} still performed well on asymmetric environment (PushT, Cube and Scene). 
We hypothesized that while the CTWM latent distance is symmetric by construction, the action-conditioned predictor is not, and it restricts planning to states actually reachable within the horizon $H$. 
The symmetric distance could therefore create local minima around irreversible transitions, while the planner  could escape them as long as $H$ is long enough to contain an alternative path, which will be interesting to investigate in more detail in future work.  

Given the high success rate on several benchmarks, we repeated planning experiments in which we further increased the goal distance from~25 to~100 steps, with a maximum planning budget of 125 steps (Table~\ref{app_table:numerical_results}). 
\ac{CTWM} also outperformed \ac{LeWM} across environments in this harder setting, and the margin widened considerably. For example, LeWM collapses on Two-Room ($17.3\%$), PushT ($15.3\%$) and Pointmaze ($28.2\%$), whereas CTWM retains near-ceiling performance on Two-Room ($98.3\%$), Pointmaze ($98.8\%$) and Reacher ($98.2\%$), while success rate on PushT decreased for both algorithms. 
These results suggest that the benefits of a commute-time-preserving geometry become more pronounced as the goal moves further away.
Thus, CTWM provides better task-agnostic representations for planning in continuous environments, especially when faced with bottlenecks and higher-dimensional state spaces (see Table~\ref{app_table:numerical_results}).
Ultimately, we confirmed that these advantages also translate to gradient-based planning strategies (see Appendix ~\ref{app:firstorder}).

\begin{table}[htb]
    \centering
    \small
    \begin{tabular}{l
        S[table-format=3.1(2)]
        S[table-format=3.1(2)]
        S[table-format=3.1(2)]
        S[table-format=3.1(2)]
        S[table-format=3.1(2)]
        S[table-format=3.1(2)]
    }
        \toprule
        Model & {Reacher} & {Cube} & {Two-Room} & {PushT} & {Scene} & {Pointmaze} \\
        \midrule
        Random        & 13.0\pm2.3 & 45.2\pm0.9 & 24.3\pm2.7  & 4.3\pm2.1  & 49.8\pm3.1 & 14.3\pm1.7  \\
        \ac{LeWM}     & 83.3\pm2.1 & 64.8\pm1.3 & 88.8\pm1.6  & 91.5\pm1.1 & 69.3\pm2.6 & 68.5\pm1.8  \\
        \ac{CTWM}     & 86.7\pm1.6 & 90.7\pm1.7 & 100.0\pm0.0 & 89.3\pm0.6 & 78.0\pm2.5 & 100.0\pm0.0 \\
        DINO-WM     & 79.3\pm5.7 & 86.0\pm5.3 & 100.0\pm0.0 & 74.7\pm4.1 & {--} & {--} \\
        PLDM    & 78.7\pm5.8 & 65.3\pm3.1 & 97.3\pm1.2 & 78.0\pm5.0 & {--} & {--} \\
        \ac{LeWM}-TS & 82.3\pm1.2 & 72.7\pm2.4 & 93.3\pm0.2 & 56.3\pm2.4 & 68.5\pm1.4 & 63.5\pm0.4 \\
        \midrule
        Random-100    & 7.3\pm1.9  & 45.5\pm2.5 & 1.0\pm0.4   & 0.3\pm0.2  & 24.2\pm1.7 & 11.3\pm1.0  \\
        \ac{LeWM}-100 & 81.0\pm0.8 & 61.7\pm1.4 & 17.3\pm0.8  & 15.3\pm1.4 & 40.3\pm2.5 & 28.2\pm1.2  \\
        \ac{CTWM}-100 & 98.8\pm0.2 & 74.5\pm1.5 & 97.3\pm0.6  & 26.2\pm2.1 & 58.3\pm2.1 & 99.2\pm0.5  \\
        \ac{LeWM}-TS-100 & 81.3\pm1.8 & 62.5\pm1.1 & 52.5\pm3.9 & 9.5\pm1.5 & 39.8\pm3.3 & 31.3\pm1.9 \\
        \bottomrule
    \end{tabular}
    \caption{Planning success rate in percent for each model trained on three seeds after ten epochs using \ac{CEM} for the standard setting by \citet{maes2026stable} with a goal distance of 25~steps and planning budget of 50~steps (top half) and for an extended goal distance of 100~steps with planning budget of 125~steps (bottom half). A baseline floor is given by the random policy for each setting and environment. In addition to LeWM, we report the scores of DINO-WM \citep{zhou2024dino} and PLDM \citep{sobal2026learning} as published by \citet{maes2026leworldmodel}.} 
\label{app_table:numerical_results}
\end{table}

\subsubsection{Commute-time geometry in continuous spaces}

\begin{figure}[tb]
        \centering
        \includegraphics[width=1\linewidth]{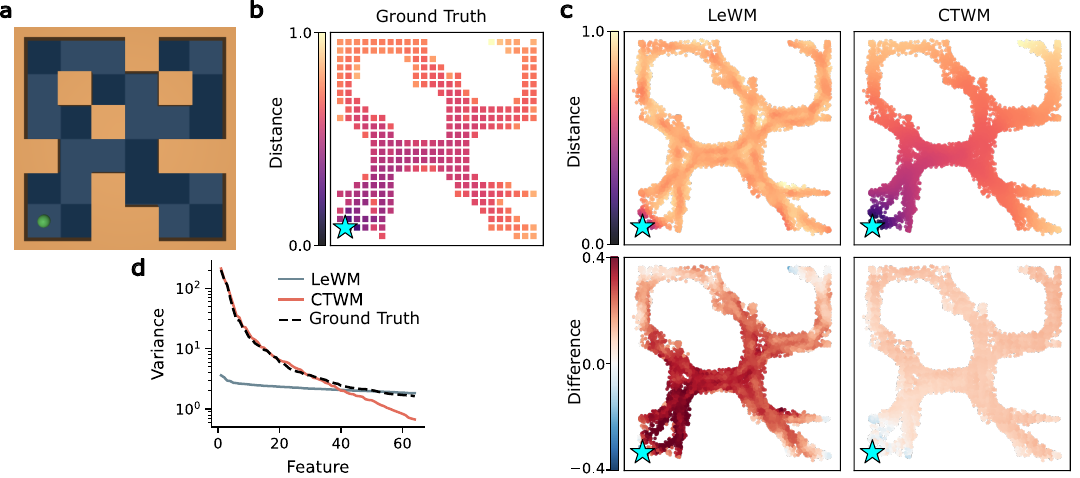}
        \caption{\textbf{Commute-time distance and variance scaling on Pointmaze.} \textbf{(a)}~The view of the Pointmaze environment. \textbf{(b)}~Ground truth commute-time distance from the blue star to any other point on the map computed from the discretized continuous state space. 
        \textbf{(c)}~Top row: Relative $L_2$ distance (normalised) in the embedding space of \ac{LeWM} (left) and \ac{CTWM} (right). 
        Bottom row: difference to the ground-truth commute-time distance. 
        \ac{CTWM} reproduces the ground-truth distance more closely. 
        \textbf{(d)}~Variance of the first 64 latent features for both models. \ac{CTWM} exhibits the $\lambda_k^{-1/2}$ decay predicted by Corollary~\ref{cor:var} (dashed line), whereas \ac{LeWM} produces a flatter spectrum.} 
        \label{fig:pointmaze}
\end{figure}

Finally, we tested whether CTWM's planning success correlates  
with its ability to faithfully represent commute-time distance in one of the above continuous environments.
To that end, we discretized the Pointmaze state space into a graph, computed the ground-truth commute-time distances from a given point (Fig.~\ref{fig:pointmaze}b), and compared them to the $L_2$ embedding space distances by \ac{LeWM} and \ac{CTWM} (Fig.~\ref{fig:pointmaze}c). 
\ac{CTWM} reproduced the ground-truth picture more faithfully. 
Then, we quantified the variance projected onto the first 64~feature dimensions (Fig.~\ref{fig:pointmaze}d). \ac{CTWM} exhibited the characteristic $\lambda_k^{-1/2}$ decay as predicted (cf.\ Corollary \ref{cor:var}), whereas \ac{LeWM} produced a flatter decay consistent with a more isotropic representation. 
Together, these results suggest that CTWM faithfully represents commute-time geometry in continuous environments learned from pixels.

\subsection{Ablations} 
\paragraph{Embedding dimension and predictor class.} For the following ablations, we report the success rate averaged over Reacher, Cube, Two-Room and Push-T. 
\ac{CTWM} and \ac{LeWM} used different embedding dimensions by default: $d = 64$ and $d = 192$, respectively. 
Thus, we compare the success rates of both models at $d = 64$ and $d = 192$. 
We also evaluated both models with an affine predictor, $T(z,a) = M(a)z + b(a)$, each at its default embedding dimension (Table \ref{table:abl_dim} and table~\ref{app_table:abl_dim}). 

\paragraph{Residual prediction and regularizer.} We compared the two regularizers, SIGReg and \ac{logdet}, each with a standard and a residual predictor, at its best embedding dimension ($d = 192$ for SIGReg and $d = 64$ for \ac{logdet}). Both residual prediction and the \ac{logdet} regularizer improved the success rate (Table~\ref{table:abl_res}, see Table~\ref{app_table:abl_res} for all results).

\begin{table}[htb]
    \centering
    \small
    \begin{subtable}[t]{0.47\linewidth}
        \centering
    \begin{tabular}{lccc}
        \toprule
        Method & $D{=}192$ & $D{=}64$ & Affine \\
        \midrule
        LeWM & $82.1\pm0.8$ & $80.6\pm0.8$ & $83.3\pm1.0$ \\
        CTWM & $\mathbf{87.8\pm0.9}$ & $\mathbf{91.6\pm0.6}$ & $\mathbf{87.6\pm0.8}$ \\
        \bottomrule
    \end{tabular}
        \caption{Embedding dimension and predictor class. ``Affine'' uses each
        model's best dimension.}
        \label{table:abl_dim}
    \end{subtable}\hfill
    \begin{subtable}[t]{0.47\linewidth}
        \centering
        \begin{tabular}{lcc}
            \toprule
            Predictor & SIGReg & Log-det \\
            \midrule
            Non-residual & $82.1\pm0.8$ & $86.4\pm1.1$ \\
            Residual & $84.7\pm0.7$ & $\mathbf{91.6\pm0.6}$ \\
            \bottomrule
        \end{tabular}
        \caption{Regularizer and residual prediction, ($d{=}192$ for SIGReg, $d{=}64$ for log-det).}
        \label{table:abl_res}
    \end{subtable}
    \caption{Ablations. Success rate (\%) averaged over Reacher, Cube, Two-Room and
    Push-T over 3 seeds. (a)~Non-affine models use a transformer predictor.
    (b)~The \ac{logdet} outperforms SIGReg with both predictors, and residual prediction
    improves both regularizers.}
    \label{table:ablations}
\end{table}

\section{Conclusion}

We introduced \ac{CTWM}, a self-supervised world model whose latent geometry is shaped for planning rather than probing.
\ac{CTWM} predicts latent displacements, regularized by the \ac{logdet} of the representation's covariance, provably recovers the scaled Laplacian representation under deterministic dynamics and at the predictor's fixed point, so that 
$L_2$ distance in latent space equals commute-time distance on the environment's transition graph. This makes the learned representation directly usable by any planner that scores candidate trajectories by 
goal distance. Across six goal-reaching tasks spanning navigation and manipulation, \ac{CTWM} matches or outperforms \ac{LeWM} while using half the parameters.
Beyond planning performance, it is striking that a purely self-supervised, reward-free objective recovers the geometry that classical graph theory identifies as optimal for planning in reversible environments. 
This raises the broader question of whether this correspondence is a quirk of our architecture, or a general principle. 
We suspect the latter. Since commute time depends only on the environment and exploration strategy, not on reward, it is a natural target for any agent that, like Tolman's rats mapping their maze before any reward is introduced \citep{tolman1948cognitive}, must build a map before it knows what that model will be used for.

\subsection*{AI use statement}

In this work, we used generative AI tools to assist with the writing of the proof in Appendix \ref{app:th1_proof}. 
and to implement parts of our experimental code.
All proofs were independently derived and verified by the authors, and all
AI-assisted code was reviewed and tested by at least two authors.
We did not use generative AI tools to generate synthetic datasets, propose or
refine hypotheses, design our research methodology or experiments, clean or
reformat datasets, or interpret results; the remaining tasks requiring
disclosure are not applicable to this work.
Additionally, we used generative AI tools to improve the readability of
author-written text, to identify related literature, and to format references.
All suggested references were manually checked against their original sources,
and all AI-assisted text was reviewed and edited by the authors.
We have reviewed all AI-assisted work and take responsibility for the final
content of this work, including text, claims or artifacts produced with the aid
of generative AI.

\subsection*{Reproducibility statement}

The code and access to datasets that we used in this paper can be found here: \url{https://github.com/fmi-basel/commute-time-preserving-world-models}.
The assumptions and complete proofs of our theoretical results are given in Appendices~\ref{app:scalings} and~\ref{app:th1_proof}.

\subsubsection*{Acknowledgments}
The authors thank Ashena Gorgan Mohammadi and all members of the Zenke Lab for their thoughtful input. This project was supported by the Swiss National Science Foundation (Grant Number PCEFP3\_202981) and the Novartis Research Foundation.

\bibliography{iclr2027_conference}
\bibliographystyle{preprintbiblio}

\newpage
\appendix
\section{Hyperparameters}
\label{app:hyper}
\begin{table}[tbh]
\centering
\small
\begin{tabular}{@{}llr@{}}
\toprule
& Hyperparameter & Value \\
\midrule
\multirow{2}{*}{Architecture}
& encoder $\phi$            & ViT-Tiny (5.5\,M) \\
& predictor $T$             & Transformer architecture of \citep{maes2026leworldmodel} of width D (3.3 M)\\
\midrule
\multirow{5}{*}{Data}
& image size                & $224\times224$ \\
& context length $H$        & 3 \\
& frame skip $k$            & 5 \\
& prediction steps          & 1 \\
& embedding dimension $D$   & {\color{hyper}64} \\
\midrule
\multirow{4}{*}{Optimisation}
& optimiser                 & AdamW \\
& learning rate             & $5\times10^{-5}$ \\
& batch size                & 128 \\
& epochs                    & 10 \\
\midrule
\multirow{4}{*}{Loss}
& $\beta$        &  {\color{hyper}$0.1$}\\
& $\gamma$        &  {\color{hyper}$1.0$}\\
& $\epsilon$        &  {\color{hyper}$10^{-6}$}\\
& chunk size        &  {\color{hyper}$32$}\\
\midrule
\multirow{5}{*}{Planning}
& solver                    & CEM: 300 samples, 30 iterations, 30 elites, $\sigma$ scale 1.0 \\
& horizon                   & 5 blocks of $k=5$ actions \\
& replanning interval       & 5 blocks \\
& goal offset / step budget & 25 / 50 environment steps \\
& evaluation episodes       & {\color{hyper}200} \\
\bottomrule
\end{tabular}
\caption{Main hyperparameters. ({\color{hyper}Green}) Changes from LeWM.}
\end{table}

\section{Regularizers}
\subsection{Orthonormality}
\label{app:ortho}
\citet{wu2018laplacian} constrain the coordinates of $\phi$ to be
orthonormal under the data distribution, enforced by the soft penalty
\begin{equation}
    Reg(\phi)
    = \frac{\delta}{2} \Bigl(
        \Bigl\lVert \tfrac{1}{N} Z^\top Z - I_d \Bigr\rVert_F^2
      + \Bigl\lVert \tfrac{1}{N} Z'^\top Z' - I_d \Bigr\rVert_F^2
      \Bigr),
\end{equation}
which drives the second moments of the representations toward the identity.

\subsection{VCReg}
\label{app:vcreg}
\ac{VICReg} \citep{bardes2021vicreg} is a common \ac{SSL} regularization methods and also motivates the loss of \citet{sobal2026learning}. 
The variance term $\mathcal{L}_v(Z, Z') =
v(Z) + v(Z')$ with
\begin{equation}
    v(Z) = \frac{1}{d} \sum_{j=1}^{d}
    \max\Bigl(0,\; \tau - \sqrt{\operatorname{Var}(z^j) + \epsilon}\Bigr)
\end{equation}
pushes the standard deviation of every coordinate above the threshold $\tau$. 
The covariance term $\mathcal{L}_c(Z, Z') = c(Z) + c(Z')$ with $c(Z) = \frac{1}{d} \sum_{i \neq j} \operatorname{Cov}(Z)_{ij}^2$ decorrelates the coordinates. 
We use only these two terms, i.e. VCReg \citep{zhu2023variance} and do not consider the invariance term as this could be seen to be implemented through the main loss:
\begin{equation}
    Reg(\phi) = \alpha_1 \mathcal{L}_c(Z, Z')
              + \alpha_2 \mathcal{L}_v(Z, Z'),
\end{equation}
which can be seen as a relaxation of the orthonormality constraint: coordinates are decorrelated and each standard deviation must reach
$\tau$, but is otherwise free.

\section{Proofs}

\subsection{Convergence of equation \ref{equ:CTWM}}
\label{app:conv}

\begin{lemma}
    \label{lemma}
    If learning achieves the optimum of the \ac{CTWM} loss function (\eqref{equ:CTWM})
\begin{equation}
    \mathcal{L}(\phi,T) = \mathbb{E}_{(s,a,s')}[\lVert\Delta(z,z') - T(z,a)\rVert^2 + \beta\lVert T(z,a)\rVert^2] + Reg(\phi),
\end{equation}
and if the dynamics are deterministic and regularization $Reg(\phi)$ is an orthonormality constraint, then the latent states $z_t$ converge to the Laplacian representation.
\end{lemma}
\begin{proof}
For simplicity, let $\Delta := z'-z$, and define the conditional mean displacement as $\mu := \mathbb{E}[\Delta|z,a] = \mathbb{E}[z'|z,a] - z$. In the following we also drop the arguments of the predictor $T(z,a)$.
Using the multivariate bias-variance decomposition
$$\mathbb{E}\left[\lVert\Delta-T\rVert^2|z,a\right] = \operatorname{tr}\bigl(\operatorname{Cov}(\Delta|z,a)\bigr) + \lVert\mu-T\rVert^2$$
the conditional objective becomes
\begin{equation}
\mathcal{L}(\phi,T)
 =  \operatorname{tr}\bigl(\operatorname{Cov}(\Delta|z,a)\bigr) + \lVert\mu-T\rVert^2 + \beta\lVert T\rVert^2 + Reg(\phi).
\end{equation}
For $\beta\geq 0$, this objective is strictly convex in $T$. Its gradient is $\nabla_T \mathcal{L}(\phi,T) = 2(T-\mu) + 2\beta T$, so the unique minimizer satisfies $\nabla_T \mathcal{L}(\phi,T)=0$. Therefore,
\begin{equation}
\label{eq:opt_pred}
T^\star(s,a) = \frac{\mu(z,a)}{1+\beta}.
\end{equation}
Substituting $T^\star$ into the objective gives
\begin{align*}
\mathcal{L}(\phi,T^\star)
    &= \mathbb{E}_{(s,a)}\left[\operatorname{tr}\bigl(\operatorname{Cov}(\Delta|z,a)\bigr) + \left\lVert\mu-\frac{\mu}{1+\beta}\right\rVert^2 + \beta\left\lVert\frac{\mu}{1+\beta}\right\rVert^2\right] + Reg\\
    &= \mathbb{E}_{(s,a)}\left[\operatorname{tr}\bigl(\operatorname{Cov}(\Delta|z,a)\bigr)\right]+ \frac{\beta^2}{(1+\beta)^2}\mathbb{E}_{(s,a)}[\lVert\mu\rVert^2]+ \frac{\beta}{(1+\beta)^2}\mathbb{E}_{(s,a)}[\lVert\mu\rVert^2]+ Reg\\
    &= \mathbb{E}_{(s,a)}\left[\operatorname{tr}\bigl(\operatorname{Cov}(\Delta|z,a)\bigr)\right]+ \frac{\beta(1+\beta)}{(1+\beta)^2}\mathbb{E}_{(s,a)}[\lVert\mu\rVert^2]+ Reg\\
&= \mathbb{E}_{(s,a)}\left[\operatorname{tr}\bigl(\operatorname{Cov}(\Delta|z,a)\bigr)\right]+ \frac{\beta}{1+\beta}\mathbb{E}_{(s,a)}[\lVert\mu(z,a)\rVert^2]+ Reg.
\end{align*}
Thus, the final objective can be written as
\begin{equation}
\boxed{
\mathcal{L}(\phi,T^\star)
=\mathbb{E}_{(s,a)}\left[\operatorname{tr}\bigl(\operatorname{Cov}(\Delta|z,a)\bigr)\right]+\frac{\beta}{1+\beta}\mathbb{E}_{(s,a)}[\lVert\mu(z,a)\rVert^2]+ Reg.
}
\end{equation}

If the dynamics of the underlying graph is deterministic, the covariance is 0 and the conditional mean displacement $\mu(z,a) = \Delta$, the objective simplifies to:
\begin{equation}
\label{equ:general_deter}
    \mathcal{L}(\phi,T^\star) = \frac{\beta}{1+\beta}\mathbb{E}_{(s,s')}[\lVert\phi(s')-\phi(s)\rVert^2] + Reg.
\end{equation}

If the regularization implements an orthonormality constraint, this is the graph drawing objective of \citet{koren2003spectral,wu2018laplacian}. As proven previously, learned representations under this loss converge to the Laplacian representation up to a rotation \citep{wang2021towards, gomez2024proper}. 
\end{proof}

\subsection{Scaling of the representation under different regularizers (heuristics)}
\label{app:scalings}

In the following we derive heuristics for the scaling of representations under different regularizers, by assuming that representations align with the eigenvectors of the Laplacian---even though not all regularizers guarantee this.
We first consider the objective with an arbitrary regularizer, and later investigate orthonormality, VCReg and \ac{logdet} as three particular instances. 

Write $c := \beta/(1+\beta)$ and assume the representation is aligned with the eigenbasis, i.e., each coordinate is a scaled eigenvector,
$\phi_k = s_k u_{\pi(k)}$ for distinct indices $\pi(1), \dots, \pi(d)$, where $d = \dim \mathcal{Z}$ and $\pi$ a permutation. 
Since the eigenvectors satisfy $\mathbb{E}[u_k u_\ell] = \delta_{k\ell}$ and
$\mathbb{E}[u_k] = 0$, we have $\operatorname{var}(\phi_k) = s_k^2$ and $\Phi = \operatorname{diag}(s_1^2, \dots, s_d^2)$. 
Applying \citet[Prop.~1]{von2007tutorial} to each coordinate and by the eigenvalue problem $L_\pi u_\ell = \lambda_\ell D
u_\ell$, the predictive term can be rewritten as $ s_ks_lu_k^\top L_\pi u_\ell = s_ks_l\lambda_\ell \delta_{k\ell}$ which implies:
\begin{equation}
  \frac{\beta}{1+\beta}\,
  \mathbb{E}_{(s,s')}\bigl[\lVert \phi(s') - \phi(s)\rVert^2\bigr]
  = 2c \sum_{k=1}^{d} \lambda_{\pi(k)}\, s_k^2 ,
  \label{eq:common}
\end{equation}
so the objective separates across directions: each contributes a cost $2c\lambda_{\pi(k)} s_k^2$, increasing in both the eigenvalue and the variance, against whatever the regularizer contributes. 
The regularizer alone therefore determines the scaling $s_k$. 
In each case, the selected eigendirections are the $d$ slowest, $\{\pi(1), \dots, \pi(d)\} = \{1, \dots, d\}$, since \eqref{eq:common} is increasing in $\lambda$.

\paragraph{Orthonormality.}
The constraint $\Phi = I_d$ fixes $s_k = 1$ directly:
\begin{equation}
  \operatorname{var}(\phi^\star_k) = 1 .
\end{equation}
The variance is the same in every direction, and the representation is the unscaled Laplacian representation $(u_k)_{k=1}^{d}$. The general setup is studied in \citet{koren2003spectral}.

\paragraph{VCReg.}
In this case, the objective is $\sum_{k=1}^{d} \bigl[2c\lambda_k s_k^2
+ \alpha\max(0, \tau - s_k)\bigr]$. 
For $s_k > \tau$ the solution is $\tau$.  
For $s_k < \tau$ the derivative is $4c\lambda_k s_k - \alpha$, which gives a minimum at $s_k = \alpha/(4c\lambda_k)$. 
In other words,
\begin{equation}
  \operatorname{var}(\phi^\star_k) =
  \begin{cases}
    \tau^2, & \lambda_k \le \dfrac{\alpha}{4c\tau}, \\[2ex]
    \dfrac{\alpha^2}{16c^2 \lambda_k^2}, & \lambda_k > \dfrac{\alpha}{4c\tau},
  \end{cases}
\end{equation}
Fast directions decays as $\lambda_k^{-2}$ whereas the slow ones are constant. Note that \citet{balestriero2022contrastive} introduce a surrogate coinciding with the soft penalty of Appendix~\ref{app:ortho}, which does not recover the commute-time scaling either.

\paragraph{Log-det.}
For the \ac{logdet} variant, the objective is $\sum_{k=1}^{d} \bigl[2c\lambda_k s_k^2 - \gamma \log s_k\bigr]$.
The derivative for each term is $4c\lambda_k s_k - \gamma/s_k$, so the unique minimum is at
\begin{equation}
  \operatorname{var}(\phi^\star_k)
  = \frac{\gamma}{4c\lambda_k}
  = \frac{\gamma(1+\beta)}{4\beta\,\lambda_k} .
\end{equation}
which is the scaling ($\propto \lambda_k^{-1}$
\eqref{eq:scaled_lap}) we are interested in for the scaled Laplacian representation. A stronger result is provided in the next section for the \ac{logdet} where the assumption of alignement with the eigenbasis is removed.

\subsection{Proof of theorem \ref{thm:scaled_lap}}
\label{app:th1_proof}

\begingroup
\def\thetheorem{\ref{thm:scaled_lap} (Restated)}
\begin{theorem}
Let $(\lambda_k,u_k)_{k\geq 0}$ denote the eigenpairs of the generalized eigenproblem of $L_\pi$, ordered as
\[
0=\lambda_0 < \lambda_1 \leq \cdots \leq \lambda_{n-1},
\]
and assume an eigengap $\lambda_d < \lambda_{d+1}$, where $d=\dim\mathcal Z$. Let
\[
\psi(s) =
\begin{pmatrix}
\lambda_1^{-1/2}u_1(s)\\
\vdots\\
\lambda_d^{-1/2}u_d(s)
\end{pmatrix}
\]
be the rank-$d$ scaled Laplacian representation.

Under deterministic dynamics, consider the reduced CTWM objective following Lemma \ref{lemma}
\begin{equation}
\label{eq:deter_app2}
\mathcal L(\phi) = \frac{\beta}{1+\beta} \mathbb E_{(s,s')} \left[ \|\phi(s')-\phi(s)\|_2^2 \right] - \frac{\gamma}{2} \log\det\Sigma, \qquad \Sigma_\phi := \operatorname{Cov}_\rho[\phi(s)],
\end{equation}

where $\rho$ is the stationary state distribution.
Then every minimizer is, up to an orthogonal transformation and a translation,
\[
\phi^\star(s) = \sqrt{\frac{\gamma(1+\beta)}{4\beta}}\, O\psi(s) +b, \qquad O\in\mathsf O(d).
\]
\end{theorem}

\addtocounter{theorem}{-1}
\endgroup

\begin{proof}
The proof proceeds in four steps. 
We first reduce the problem to a matrix optimization (Step~1), then factor the coefficient matrix into the subspace $R$ spanned by the features and the covariance $\Sigma$ that scales it, rewriting the loss in terms of both (Step~2). 
We then minimize over $\Sigma$ for a fixed subspace (Step~3), and substituting the result back leaves a problem in $R$ alone, which we solve with the Poincaré separation theorem (Step~4). 

\paragraph{Notation.} Define $c := \beta/(1+\beta)$.
Since both terms in the objective are invariant under translation of the representation, we may assume without loss of generality that $ \mathbb E_\rho[\phi(s)]=0 $.

Let $u_1,\ldots,u_{n-1}$ denote the nonconstant Laplacian eigenfunctions, normalized with respect to the stationary state distribution $\rho$ such that $ \mathbb E_\rho[u_k(s)u_\ell(s)] = \delta_{k\ell} $.
Collect them in
\[
u(s) =
\begin{pmatrix}
u_1(s)\\
\vdots\\
u_{n-1}(s)
\end{pmatrix},
\qquad \Lambda = \operatorname{diag} (\lambda_1,\ldots,\lambda_{n-1}).
\]

Because the $u_k$ form an orthonormal basis of the centered functions on the finite state space, any centered $d$-dimensional representation can be expanded as
\[
\phi(s)=A u(s), \qquad A\in\mathbb R^{d\times(n-1)}.
\]

\paragraph{Step 1: Reduction to a matrix optimization.}

By orthonormality of the Laplacian eigenfunctions,
\[
\Sigma = \operatorname{Cov}_\rho[\phi(s)] = AA^\top.
\]

We will now rewrite \eqref{eq:deter_app2} in terms of $A$. We first apply to its first term \citet[Prop.~1]{von2007tutorial} to each coordinate: 
\begin{equation*}
    \mathbb{E}\lVert \phi(s') - \phi(s)\rVert^2 = \sum_j \mathbb{E}(\phi_j(s') - \phi_j(s))^2 = 2\sum_j\phi_j^\top L_\pi \phi_j.
\end{equation*} 
In addition, if we use the definition of the generalized eigenproblem $L_\pi u_\ell = \lambda_\ell D u_\ell$, we have:
\begin{equation*}
    u_k^\top L_\pi u_\ell = \lambda_\ell u_k^\top
    D u_\ell = \lambda_\ell \mathbb{E}[u_k(s) u_\ell(s)]= \lambda_\ell \delta_{k\ell}.
\end{equation*} 
Combining the two last equations leads to:
\begin{equation*}
    \mathbb{E}\lVert \phi(s') - \phi(s)\rVert^2 = 2\sum_j\sum_{k,\ell} A_{jk} A_{j\ell}\, u_k^\top L_\pi\, u_\ell = 2\sum_j\sum_k \lambda_k A_{jk}^2 =  2 \operatorname{tr}(A \Lambda A^\top).
\end{equation*}
with
$\Lambda = \operatorname{diag}(\lambda_1, \dots, \lambda_{n-1})$. The
objective therefore becomes
\begin{equation}
\label{eq:matrix_objective}
  \mathcal{L}
  = 2c \operatorname{tr}(A \Lambda A^\top)
    - \tfrac{\gamma}{2} \log \det (A A^\top).
\end{equation}

\paragraph{Step 2: Separate the represented subspace from its scaling.}

Let
\[
\Sigma:=AA^\top\succ 0,
\qquad
R:=\Sigma^{-1/2}A.
\]
Then
\[
RR^\top
=
\Sigma^{-1/2}AA^\top\Sigma^{-1/2}
=
I_d,
\]
and
\[
A=\Sigma^{1/2}R.
\]

This decomposition separates two properties of the representation:
$R$ determines the $d$-dimensional subspace represented by $\phi$,
whereas $\Sigma$ determines the covariance, and hence the scaling,
within that subspace.

Substituting $A=\Sigma^{1/2}R$ into
\eqref{eq:matrix_objective} gives
\begin{align}
\mathcal L(\Sigma,R)
&=
2c\,
\operatorname{tr}
\left(
\Sigma^{1/2}R\Lambda R^\top\Sigma^{1/2}
\right)
-
\frac{\gamma}{2}\log\det\Sigma
\\
&=
2c\,
\operatorname{tr}
\left(
\Sigma R\Lambda R^\top
\right)
-
\frac{\gamma}{2}\log\det\Sigma.
\label{eq:cov_subspace_objective}
\end{align}

Define the Laplacian restricted to the selected subspace as
\[
M_R:=R\Lambda R^\top.
\]
Since $\lambda_k>0$ for all $k\geq1$ and $R$ has full row rank,
$M_R\succ0$.

\paragraph{Step 3: Optimal covariance for a fixed subspace.}

For fixed $R$, \eqref{eq:cov_subspace_objective} is minimized
over $\Sigma\succ0$.
Using
\[
\nabla_\Sigma \operatorname{tr}(\Sigma M_R)=M_R,
\qquad
\nabla_\Sigma\log\det\Sigma=\Sigma^{-1},
\]
the stationary condition is
\[
2cM_R-\frac{\gamma}{2}\Sigma^{-1}=0.
\]
Therefore,
\begin{equation}
\Sigma_R^\star
=
\frac{\gamma}{4c}
M_R^{-1}
=
\frac{\gamma}{4c}
\left(R\Lambda R^\top\right)^{-1}.
\label{eq:optimal_covariance}
\end{equation}

Thus, for any fixed spectral subspace, the optimal representation
covariance is proportional to the inverse Laplacian restricted to
that subspace.

Substituting \eqref{eq:optimal_covariance} back into the
objective gives
\begin{align*}
\mathcal L(\Sigma_R^\star,R)
&=
2c\,
\operatorname{tr}
\left(
\frac{\gamma}{4c}
M_R^{-1}M_R
\right)
-
\frac{\gamma}{2}
\log\det
\left(
\frac{\gamma}{4c}M_R^{-1}
\right)
\\
&=
\frac{\gamma d}{2}
-
\frac{\gamma d}{2}
\log\frac{\gamma}{4c}
+
\frac{\gamma}{2}\log\det M_R.
\end{align*}
The first two terms do not depend on $R$. Hence optimizing the
subspace is equivalent to
\begin{equation*}
\min_{RR^\top=I_d}
\log\det(R\Lambda R^\top).
\end{equation*}

\paragraph{Step 4: The optimal subspace contains the slowest modes.}

Let
\[
\mu_1\leq\cdots\leq\mu_d
\]
denote the eigenvalues of $R\Lambda R^\top$.
By the Poincar\'e separation theorem,
\[
\lambda_k\leq\mu_k,
\qquad
k=1,\ldots,d.
\]
Therefore,
\[
\det(R\Lambda R^\top)
=
\prod_{k=1}^d \mu_k
\geq
\prod_{k=1}^d \lambda_k.
\]

Equality is attained when the row space of $R$ is the span of the
first $d$ Laplacian eigendirections. Because
$\lambda_d<\lambda_{d+1}$, this subspace is unique. Hence, up to an
orthogonal transformation $O\in\mathsf O(d)$,
\[
R^\star
=
O
\begin{bmatrix}
I_d & 0
\end{bmatrix}.
\]

Writing
\[
\Lambda_d
=
\operatorname{diag}(\lambda_1,\ldots,\lambda_d),
\]
we consequently have
\[
M_{R^\star}
=
R^\star\Lambda R^{\star\top}
=
O\Lambda_dO^\top.
\]
\Eqref{eq:optimal_covariance} then gives
\[
\Sigma^\star
=
\frac{\gamma}{4c}
O\Lambda_d^{-1}O^\top.
\]

One corresponding coefficient matrix is
\[
A^\star
=
\sqrt{\frac{\gamma}{4c}}\,
O
\begin{bmatrix}
\Lambda_d^{-1/2} & 0
\end{bmatrix}.
\]
Thus
\begin{align*}
\phi^\star(s)
&=
A^\star u(s) =
\sqrt{\frac{\gamma}{4c}}\,
O \psi(s).
\end{align*}

Finally, substituting
$c=\beta/(1+\beta)$ yields
\[
\phi^\star(s)
=
\sqrt{
\frac{\gamma(1+\beta)}{4\beta}
}\,
O\psi(s).
\]
Restoring the arbitrary translation $b$ gives
\[
\phi^\star(s)
=
b+
\sqrt{
\frac{\gamma(1+\beta)}{4\beta}
}\,
O\psi(s)
\]

which corresponds to \eqref{eq:opt_phi}.
\end{proof}

\subsection{Proof of Corollary \ref{cor:var}}
\label{app:corollary}
\begin{proof}
By definition of the eigenvectors $(u_k)_{k>0}$ of the Laplacian $\mathbb{E}[u_k(s) u_\ell(s)] =
\delta_{k\ell}$ \citep{belkin2001laplacian}. Moreover $u_0 \equiv 1$ is itself an eigenvector, so orthogonality to it gives $\mathbb{E}[u_k] = \mathbb{E}[u_k \cdot 1] = 0$ for $k \ge 1$. Hence $\operatorname{Cov}(u_k, u_\ell) = \mathbb{E}[u_k(s) u_\ell(s)] - \mathbb{E}[u_k(s)]\mathbb{E}[u_\ell(s)] =
\delta_{k\ell}$, and since $\psi_k = \lambda_k^{-1/2} u_k$,
\begin{equation}
  \operatorname{Cov}(\phi^\star) = \frac{\gamma(1+\beta)}{4\beta}\operatorname{Cov}(\psi)
  = \frac{\gamma(1+\beta)}{4\beta}\Lambda_d^{-1}.
\end{equation}
\end{proof}

\section{Smoothness constraint}

\subsection{Smoothness with next-state prediction}
\label{app:smoothness}

Instead of using residual prediction as in \ref{equ:CTWM}, we could reparametrize it into a mathematically equivalent next-state prediction loss. Let's denote with $N(\phi(s), a)$ the network to predict the next state embedding starting form the current state embedding and the action. We can write the following loss:
\begin{equation}
    \mathcal{L}(\phi,T) = \mathbb{E}_{(s,a,s')}[\lVert \phi(s') - N(\phi(s),a)\rVert^2 + \beta\lVert N(\phi(s),a)-\phi(s)\rVert^2] + Reg(\phi).
\end{equation}
Now, if we substitute $N(\phi(s), a)=\phi(s) + T(\phi(s), a)$ we get:
\begin{equation*}
    \mathcal{L}(\phi,T) = \mathbb{E}_{(s,a,s')}[\lVert\phi(s') - \phi(s)- T(\phi(s),a)\rVert^2 + \beta\lVert \phi(s) + T(\phi(s), a)-\phi(s)\rVert^2] + Reg(\phi).
\end{equation*}
By simplifying we obtain:
\begin{equation}
    \mathcal{L}(\phi,T) = \mathbb{E}_{(s,a,s')}[\lVert\phi(s') - \phi(s)- T(\phi(s),a)\rVert^2 + \beta\lVert T(\phi(s), a)\rVert^2] + Reg(\phi).
\end{equation}
Eventually, denoting with $\Delta(s,s') = \phi(s')-\phi(s)$, we have:
\begin{equation}
    \mathcal{L}(\phi,T) = \mathbb{E}_{(s,a,s')}[\lVert\Delta(s,s')- T(\phi(s),a)\rVert^2 + \beta\lVert T(\phi(s), a)\rVert^2] + Reg(\phi).
\end{equation}
This is the equation \ref{equ:CTWM}.\\

\subsection{Planning with beta}
\label{app:beta}
The fixed point condition (\eqref{eq:opt_pred}) implies that our planning method should rely on:
\begin{equation}
    \hat{z}_{t+1} = z + (1+\beta)T(z,a_t).
\end{equation}
We trained the model with different beta values and estimated the $\beta_\mathrm{est}$ that would minimize the scaling error from the predictor for each of them using Ordinary Least-Squares (OLS):
\begin{equation}
     z_{t+1}- z_t = (1+\beta_\mathrm{est})T(z,a_t) \quad .  
\end{equation}
The results (Table \ref{tab:beta}) show that the estimated scaling for the predictor is close to the one computed in equation \ref{eq:opt_pred} on the PushT environment.
\begin{table}[!htp]
    \centering
    \small
    \begin{tabular}{lcc}
        \toprule
         $\beta$ &  $\beta_\mathrm{est}$ \\
        \midrule
        0.005 & 0.0074\\
        0.1 & 0.108\\
        0.25 & 0.26\\
        0.5 & 0.52\\
        1.0 & 1.04\\
        \bottomrule
    \end{tabular}
    \caption{Comparison of $\beta$ and $\beta_\mathrm{est}$ on PushT.}
    \label{tab:beta}
\end{table}

\section{Planning with first-order methods}
\label{app:firstorder}
\begin{table}[!htbp]
    \centering
    \small
    \begin{tabular}{l
        S[table-format=3.1(2)]
        S[table-format=3.1(2)]
        S[table-format=3.1(2)]
        S[table-format=3.1(2)]
        S[table-format=3.1(2)]
        S[table-format=3.1(2)]
    }
        \toprule
        Model & {Reacher} & {Cube} & {Two-Room} & {PushT} & {Scene} & {Pointmaze} \\
        \midrule
        Random    & 13.0\pm2.3 & 45.2\pm0.9 & 24.3\pm2.7 & 4.3\pm2.1  & 49.8\pm3.1 & 14.3\pm1.7  \\
        \ac{LeWM} & 73.7\pm2.7 & 60.2\pm0.5 & 93.7\pm2.5 & 86.5\pm1.2 & 66.7\pm2.4 & 88.3\pm1.2  \\
        \ac{CTWM} & 75.5\pm0.4 & 80.2\pm2.4 & 98.5\pm0.4 & 84.0\pm1.9 & 74.5\pm2.0 & 100.0\pm0.0 \\
        \bottomrule
    \end{tabular}
    \caption{Planning success rate in percent for each model trained on three seeds after ten epochs using Adam. A floor is given by the random policy for each environment. } 
    \label{tab:results_adam}
\end{table}

\section{Ablations}
\subsection{Rooting of smoothness constraint}
We also rooted the smoothness constraint in the encoder replacing the regularization term $||T(\phi(s),a)||^2$ by $||\phi(s)-\phi(s')||^2$. Except removing the beta during planning this does not improved performances (Table \ref{app_tab:enc_root}). 
\begin{table}[htb]
    \centering
    \begin{tabular}{lccccc}
        \toprule
        Method & Two-Room & Reacher & Push-T & Cube & Mean \\
        \midrule
        Reg. rooted in predictor   & $100.0\pm0.0$ & $86.7\pm1.5$ & $89.2\pm0.9$ & $90.7\pm1.4$ & $91.6\pm0.6$ \\
        Reg. rooted in encoder & $100.0\pm0.0$ & $84.7\pm3.3$ & $89.7\pm0.5$ & $91.3\pm0.2$ & $91.4\pm0.8$ \\
        \bottomrule
    \end{tabular}
    \caption{The smoothness constraint can be rooted in the predictor or encoder. Both results in similar performances (3 seeds, beta=0.1).}
    \label{app_tab:enc_root}
\end{table}

\newpage
\subsection{Loss and Residual}
\begin{table}[htb]
    \centering
    \small
    \begin{tabular}{llccccc}
        \toprule
        Reg. & Predictor & Two-Room & Reacher & Push-T & Cube & Mean \\
        \midrule
        \multirow{2}{*}{SIGReg ($D{=}192$)} & Non-residual & $88.8\pm1.5$ & $83.3\pm2.1$ & $\mathbf{91.5\pm1.1}$ & $64.8\pm1.3$ & $82.1\pm0.8$ \\
         & Residual & $90.3\pm0.2$ & $81.8\pm2.1$ & $88.7\pm0.6$ & $78.0\pm1.6$ & $84.7\pm0.7$ \\
        \midrule
        \multirow{2}{*}{Log-det ($D{=}64$)} & Non-residual & $99.8\pm0.2$ & $83.8\pm3.3$ & $88.2\pm2.8$ & $73.8\pm0.6$ & $86.4\pm1.1$ \\
         & Residual & $\mathbf{100.0\pm0.0}$ & $\mathbf{86.7\pm1.5}$ & $89.2\pm0.9$ & $\mathbf{90.7\pm1.4}$ & $\mathbf{91.6\pm0.6}$ \\
        \bottomrule     
    \end{tabular}
    \caption{Ablations. Success rate (\%) averaged over Reacher, Cube, Two-Room and Push-T over 3 seeds.}
    \label{app_table:abl_res}
\end{table}

\subsection{Predictor size}
\begin{table}[htb]
    \centering
    \begin{tabular}{llccccc}
        \toprule
        Method & Config. & Two-Room & Reacher & Push-T & Cube & Mean \\
        \midrule
        \multirow{3}{*}{LeWM} & $D{=}192$ & $88.8\pm1.5$ & $83.3\pm2.1$ & $\mathbf{91.5\pm1.1}$ & $64.8\pm1.3$ & $82.1\pm0.8$ \\
         & $D{=}64$ & $92.3\pm1.0$ & $83.0\pm2.3$ & $82.2\pm0.9$ & $65.0\pm1.5$ & $80.6\pm0.8$ \\
         & Affine & $89.7\pm0.2$ & $70.2\pm1.9$ & $84.0\pm2.4$ & $89.5\pm2.5$ & $83.3\pm1.0$ \\
        \midrule
        \multirow{3}{*}{CTWM} & $D{=}192$ & $99.8\pm0.2$ & $84.5\pm2.7$ & $79.0\pm2.1$ & $87.8\pm1.2$ & $87.8\pm0.9$ \\
         & $D{=}64$ & $\mathbf{100.0\pm0.0}$ & $\mathbf{86.7\pm1.5}$ & $89.2\pm0.9$ & $90.7\pm1.4$ & $\mathbf{91.6\pm0.6}$ \\
         & Affine & $100.0\pm0.0$ & $71.3\pm2.7$ & $82.5\pm1.1$ & $\mathbf{96.5\pm0.7}$ & $87.6\pm0.8$ \\
        \bottomrule
    \end{tabular}
    \caption{Ablations. Success rate (\%) averaged over Reacher, Cube, Two-Room and Push-T over 3 seeds.}
    \label{app_table:abl_dim}
\end{table}
\section{Subspace similarity}
\label{app:subsim}
        The subspace similarity is measured using the canonical angles (principal angles) method, a mathematical technique used to measure the alignment and distance between two linear subspaces.
        Let $\Phi \in \mathbb{R}^{n \times d}$ collect the learned embeddings of the $n$ graph nodes, and let $U \in \mathbb{R}^{n \times d}$ collect the Laplacian eigenvectors $u_1, \dots, u_d$, excluding the constant eigenvector $u_0$.
        We first subtract the mean embedding from $\Phi$, since the Laplacian eigenvectors are orthogonal to the constant vector and the learned embeddings are only defined up to a translation.
        We then take the left singular vectors of the centered $\Phi$ as an orthonormal basis $A$ of its column space, and similarly obtain $B$ from $U$. The singular values $\sigma_1 \geq \dots \geq \sigma_d$ of $A^\top B$ are the cosines of the principal angles between the two subspaces~\citep{bjorck1973numerical}, and we report their mean, $\frac{1}{d}\sum_i \sigma_i \in [0, 1]$, which equals~$1$ for identical and~$0$ for orthogonal subspaces. The same procedure is applied to all methods. This measure is invariant to rotations within degenerate eigenspaces, which is essential on the torus, and to any rescaling of the embedding; eigenvalue-dependent scaling is therefore assessed separately (Fig.~\ref{fig:torus}c).
\end{document}